%% file: main.tex
\documentclass[10pt,twocolumn,letterpaper]{article}

\usepackage[pagenumbers]{cvpr} 

\input{preamble}

\definecolor{cvprblue}{rgb}{0.21,0.49,0.74} 

\usepackage{bm,multirow}
\usepackage{xcolor}
\usepackage{pifont}
\usepackage[dvipsnames]{xcolor}
\usepackage{float}
\usepackage[pagebackref,breaklinks,colorlinks,allcolors=cvprblue]{hyperref}
\usepackage[accsupp]{axessibility}

\def\paperID{45152}
\def\confName{CVPR}
\def\confYear{2026}
\title{\Approach: Lightning-fast Subject-driven Image Personalization\\via One step Diffusion}

\author{
Huy Duong$^{1}$\thanks{Equal contribution} \hspace{2mm} Trong-Tung Nguyen$^{1}$\footnotemark[1] \hspace{2mm} Cuong Pham$^{1,2}$ \hspace{2mm} Anh Tran$^{1}$ \hspace{2mm} Khoi Nguyen$^{1}$ \hspace{2mm} Minh Hoai$^{1}$ \\
{\small{$^{1}${Qualcomm AI Research}\thanks{Qualcomm AI Research is an initiative of Qualcomm Technologies, Inc.} \hspace{3mm} $^{2}${Posts \& Telecommunication Institute of Technology} \hspace{3mm}}} \\
{\tt\small {\{huyduong, tunnguy, pcuong, anhtra, khoi, minhhoai\}}@qti.qualcomm.com} \\
}

\input{definitions} 

\begin{document}

\makeatletter
\g@addto@macro\@maketitle{\vspace{-13mm}
  \begin{figure}[H]
  \setlength{\linewidth}{\textwidth}
  \setlength{\hsize}{\textwidth}
  \centering
  \includegraphics[width=\textwidth]{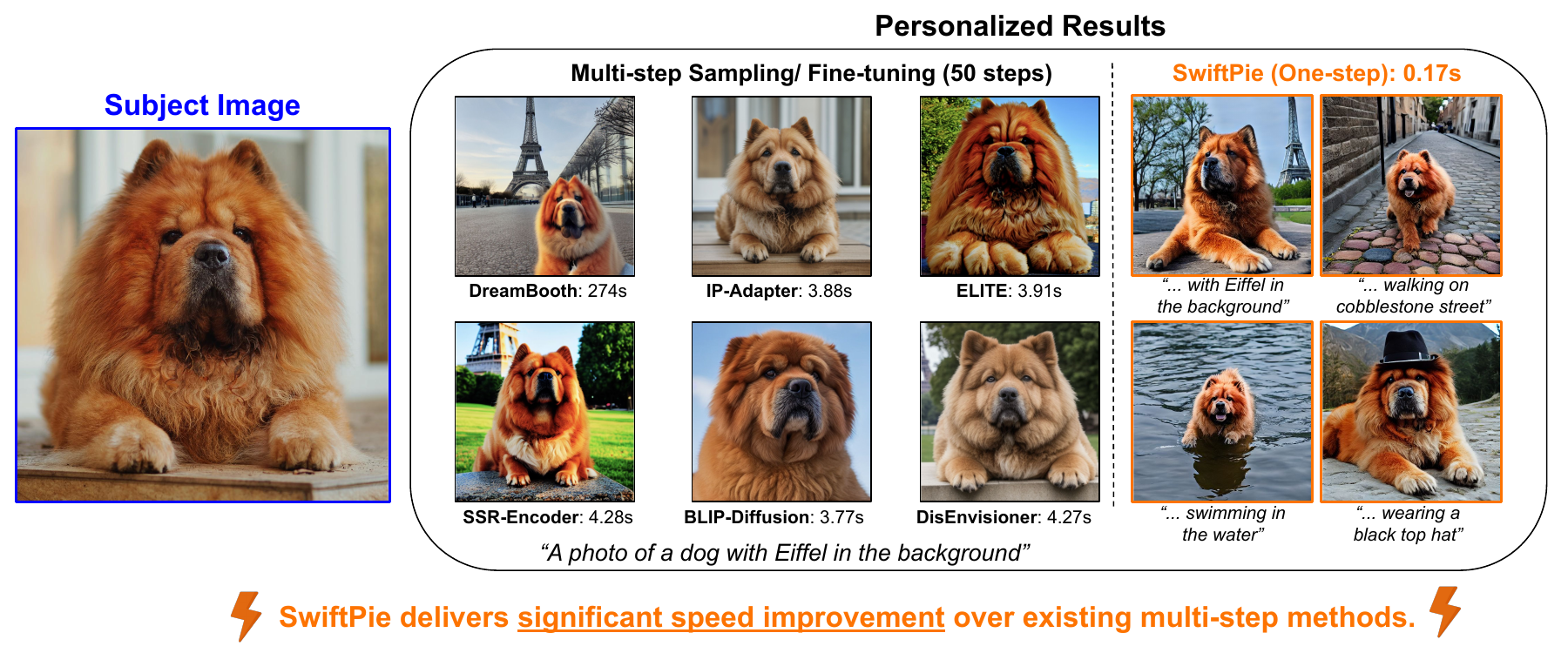}
  \vspace{-3mm}
  \caption{Given an image for a reference subject, SwiftPie generates personalized images with high-fidelity subject identity and strong text alignment \textbf{in just 0.17 seconds} on a single A100 GPU. In contrast, existing methods suffer from significantly slower generation speeds (e.g., over 4 minutes for DreamBooth), bad trade-off in subject identity preservation and text alignment.} \label{fig:teaser}
  \end{figure}
}

\maketitle
\input{sec/0_abstract}    
\input{sec/1_intro}

\input{sec/2_related_work}
\input{sec/3_preliminaries}

\input{sec/4_methods}
\input{sec/5_experiments}

\input{sec/6_ablate}
\input{sec/7_conclusion}


\clearpage
{
    \small
    \bibliographystyle{ieeenat_fullname}
    \bibliography{main}
}
\input{sec/X_suppl}


\end{document}

%% file: preamble.tex
\newcommand{\TODO}[1]{\textbf{\color{red}[TODO: #1]}}
\renewcommand{\TODO}[1]{}

\usepackage{microtype}

\renewcommand{\paragraph}[1]{\vspace{.5em}\noindent\textbf{#1.}}

\usepackage{xspace}

\usepackage{soul}
\setuldepth{foobar}

%% file: definitions.tex
\def\Approach{SwiftPie}

\def\mD{\mathcal{D}}

\def\mG{\mathcal{G}}

\def\mL{\mathcal{L}}

\def\1n{\mathbf{1}_n}
\def\0{\mathbf{0}}
\def\1{\mathbf{1}}

\def\K{{\bf K}}

\def\M{{\bf M}}

\def\Q{{\bf Q}}

\def\V{{\bf V}}
\def\W{{\bf W}}

\def\c{{\bf c}}

\def\e{{\bf e}}

\def\h{{\bf h}}

\def\s{{\bf s}}

\def\x{{\bf x}}
\def\y{{\bf y}}
\def\z{{\bf z}}

\newcommand{\minisection}[1]{\vspace{1ex}\noindent \textbf{#1}}

\newcommand{\cm}[1]{}

\newcommand{\myheading}[1]{\vspace{1ex}\noindent \textbf{#1}}

\newif\ifshowsolution
\showsolutiontrue
\ifshowsolution

\else

\fi

\newcommand{\cmark}{\textcolor{green!60!black}{\ding{51}}} 
\newcommand{\xmark}{\textcolor{red}{\ding{55}}}            

%% file: sec/0_abstract.tex
\begin{abstract}
Diffusion models have achieved remarkable success in high-quality image synthesis, sparking interest in image-guided generation tasks such as subject-driven image personalization. Despite their impressive personalization results, existing methods typically rely on computationally intensive fine-tuning, iterative optimization, or multi-step denoising processes, which significantly hinder their deployment and interactive capability in real-time applications. In this work, we present \Approach, the first one-step diffusion image personalization tool that enables lightning-fast generation of personalized images. \Approach \/ introduces a novel dual-branch identity injection mechanism that effectively integrates subject identity into a one-step diffusion model. In addition, we incorporate a mask-guided rescaling strategy to further enhance subject contextualization within a single diffusion step. Extensive experiments demonstrate that \Approach \/ not only delivers superior image personalization speed but also achieves comparable performance with multi-step approaches in both identity fidelity and prompt alignment. This work opens new opportunities for real-time, high-quality personalized image generation, paving the way for interactive visual synthesis.
\end{abstract}

%% file: sec/1_intro.tex
\section{Introduction}
\label{sec:intro}
Diffusion models have recently emerged as a cornerstone of deep generative modeling, demonstrating impressive performance in text-to-image synthesis \cite{10.5555/3600270.3602913, podell2024sdxl, Rombach_2022_CVPR} and other visual generation tasks \cite{Mokady_2023_CVPR, hertz2023prompttoprompt, cao_2023_masactrl, nguyen2024flexeditflexiblecontrollablediffusionbased, Nguyen_2025_CVPR, Avrahami_2022_CVPR, nguyen2024editscoutlocatingforgedregions}. These models typically rely on a multi-step denoising process, in which pure random noise is gradually refined into high-quality image samples \cite{song2021scorebased, song2022denoisingdiffusionimplicitmodels, 10.5555/3495724.3496298}. While this iterative mechanism is key to their success, long inference process often limits their applicability in interactive or real-time scenarios. To address this, recent works have proposed to accelerate the diffusion process by reducing the number of sampling steps, from hundreds to only a few \cite{10.1145/3680528.3687625, salimans2022progressive, Meng_2023_CVPR}, or even a single step \cite{nguyen2024swiftbrush, dao2025swiftbrush, yin2024onestep, yin2024improved}. The emergence of one-step diffusion models represents a remarkable leap forward, enabling instant image generation while opening new opportunities to extend their strong priors to downstream tasks that demand both speed and controllability. 

Such strong priors encoded in diffusion models provide rich information about how images and texts are jointly distributed, enabling users to guide image synthesis with diverse forms of controllable input---not only text but also visual clues such as reference images or masks. This capability has driven rapid progress in visual guidance generation tasks such as image editing \cite{Mokady_2023_CVPR, hertz2023prompttoprompt, cao_2023_masactrl, Nguyen_2025_CVPR, nguyen2024flexeditflexiblecontrollablediffusionbased}, inpainting \cite{Avrahami_2022_CVPR, xie2025turbofill, 10.1007/978-3-031-72661-3_9, Lugmayr_2022_CVPR}, and image personalization \cite{Ruiz_2023_CVPR, Wei2023ELITEEV, ye2023ip-adapter, he2025disenvisioner}. Building on the efficiency of few-step and one-step diffusion, recent works have adapted these fast models for controllable generation. For example, TurboEdit \cite{10.1145/3680528.3687612} and SwiftEdit \cite{Nguyen_2025_CVPR} proposed to incorporate additional visual guidance into a few-step or one-step text-to-image generation model for image editing tasks. On the other hand, TurboFill \cite{xie2025turbofill} adapted the few-step diffusion model to the image-inpainting task via additional multistage training. Despite these progresses, the extension of one-step diffusion models to subject-driven image personalization remains underexplored. Latest personalization approaches still rely on the expensive multi-step inference process, which requires some additional fine-tuning/optimization stage on new subject concepts, resulting in slow inference speed and limited user interaction. In this work, we bridge this gap by introducing the first \textbf{one-step subject-driven image generation} framework.

Developing a one-step diffusion model for subject-driven image generation, however, is a nontrivial challenge. This task requires adapting an existing one-step text-to-image diffusion model by incorporating an adapter module that injects subject identity. Yet, existing adapter-based methods, such as IP-Adapter~\cite{ye2023ip-adapter}, are not well suited for this purpose, for two main reasons. First, these adapters are primarily designed for multi-step diffusion; naively applying them in the one-step setting often leads to poor identity fidelity. Second, they tend to encode only global image features from the reference image, lacking mechanisms to preserve fine-grained subject-specific details crucial for high-fidelity personalization.

To overcome the challenges of degraded identity fidelity and insufficient fine-grained detail preservation in one-step diffusion models, we introduce a novel framework built on top of a pre-trained one-step text-to-image generator. Our key architectural innovation is a {\bf dual-branch identity injection framework}, which integrates subject-specific features into both self- and cross-attention layers. This design enables the model to retain strong identity fidelity by injecting both global and fine-grained visual cues throughout the generation process. To further enhance prompt alignment, we propose a second technique: {\bf mask-guided rescaling}, which adaptively balances subject identity injection with text-conditioned guidance during inference.

\begin{figure}
    \centering
    \includegraphics[width=\columnwidth]{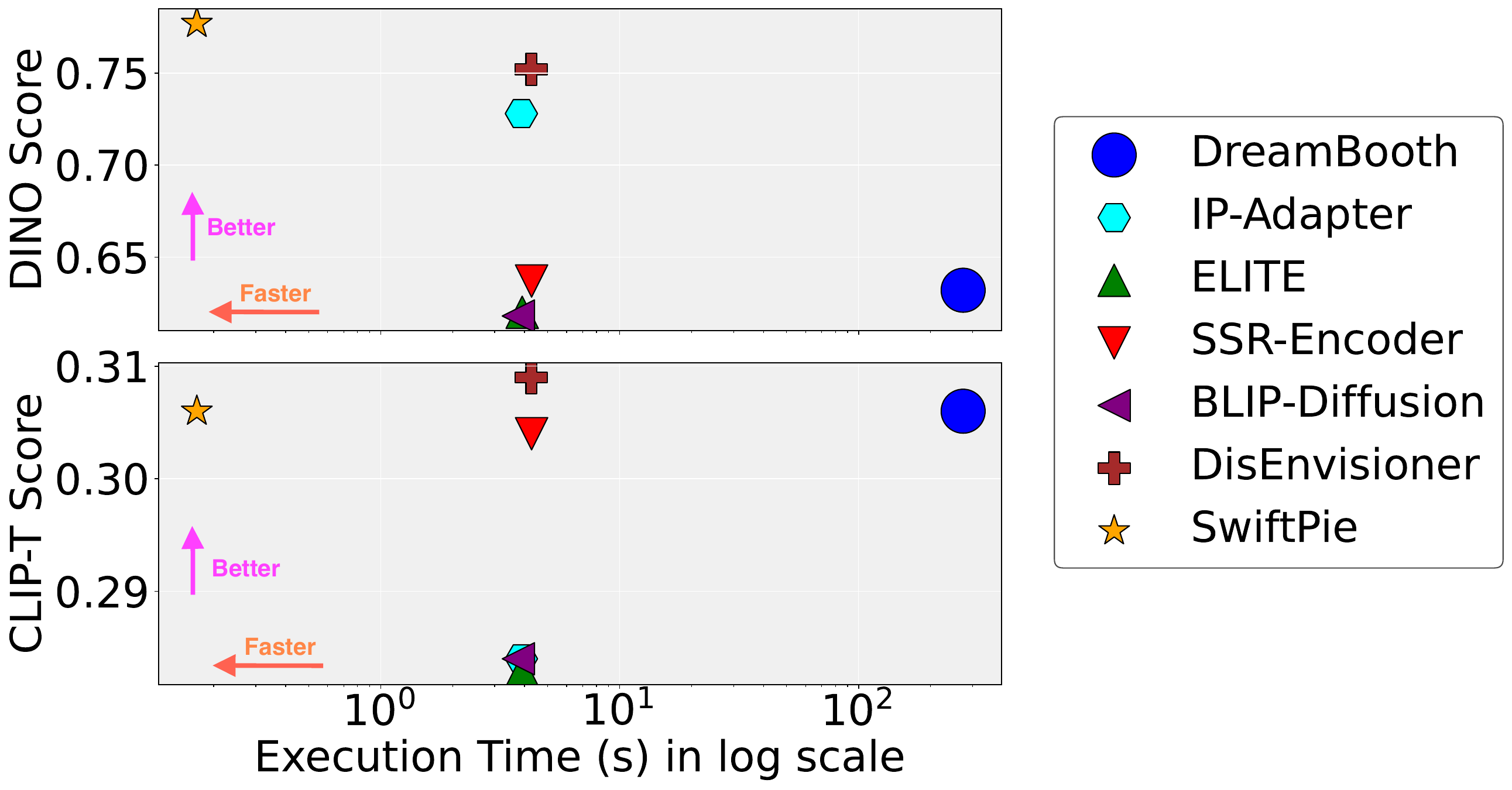}
    \caption{Performance–speed comparison of our one-step SwiftPie against multi-step personalization methods. The top-left point in each subplot denotes the optimal trade-off. SwiftPie offers the fastest speed with comparable quality.}
    \label{fig:plot_compare}
\end{figure}

Our framework achieves both fast inference and high-quality personalized outputs in the common DreamBench~\cite{Ruiz_2023_CVPR} dataset, outperforming multi-step baselines in identity fidelity, prompt alignment, and runtime efficiency. The qualitative results are shown in \cref{fig:teaser} while the quantitative results are shown in \cref{fig:plot_compare}. 
 
In summary, we are the first to propose an effective one-step diffusion framework for subject-driven image generation, achieving high-quality results with fast inference by introducing a dual-branch identity injection module that preserves global and fine-grained subject features by integrating identity cues into self- and cross-attention layers, supported by training objectives that enhance realism. For improving prompt alignment while preserving identity fidelity, we also develop a mask-guided rescaling technique.




%% file: sec/2_related_work.tex
\section{Related Work}
\label{sec:related_works}

\begin{figure*}[t]
    \centering
    \includegraphics[width=0.9\textwidth]{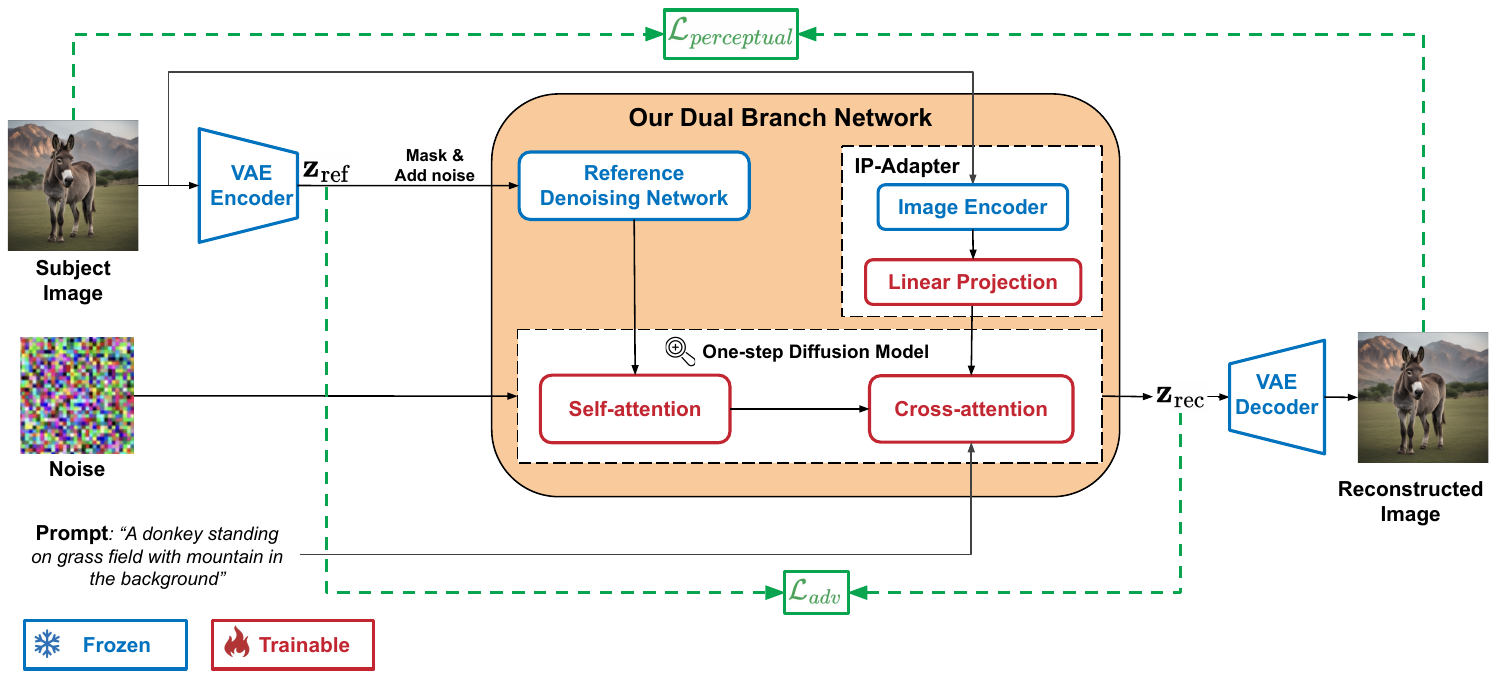}
    \caption{\textbf{Training framework for dual-branch identity injection.} Subject identity features are injected through two pathways: the Reference Network captures fine-grained features via self-attention, while the IP-Adapter encodes coarse features via cross-attention. We employ two weak reconstruction objectives to enable diverse yet identity-preserving images: a perceptual loss in image space and an adversarial loss in latent space.}
    \label{fig:main_diagram}
    \vspace{-10pt}
\end{figure*}

\subsection{Text-to-image Diffusion Models}
Text-to-image diffusion models \cite{Rombach_2022_CVPR, 10.1145/3680528.3687625, 10.5555/3600270.3602913} typically rely on a multi-step denoising diffusion process, starting from random Gaussian noise and gradually refining it into a clean image latent. This often requires dozens to hundreds of sampling steps for high-quality image generation, which is computationally expensive and hinders their application to interactive content generation. Recent advances have proposed accelerating this sampling process by distilling knowledge from multi-step models into few-step models \cite{10.1145/3680528.3687625, salimans2022progressive, Meng_2023_CVPR}. Notably, distillation can also be further extended to produce one-step models \cite{nguyen2024swiftbrush, dao2025swiftbrush, yin2024onestep, yin2024improved}, demonstrating remarkable performance in generating high-quality images in a single diffusion step. For example, Instaflow \cite{liu2023instaflow} employs rectified flow to train a one-step network, while DMD \cite{yin2024onestep, yin2024improved} leverages distribution-matching training objectives for effective knowledge transfer to a one-step model. SwiftBrush \cite{nguyen2024swiftbrush, dao2025swiftbrush} is an image-free distillation method that used text-to-3D generation objectives, i.e., SDS \cite{poole2023dreamfusion} and VSD \cite{10.5555/3666122.3666490}, along with additional post-training model merging and clamped CLIP loss to further enhance text alignment. 
Despite the advantages of one-step generation and recent progress in building such models, their extension to subject-driven personalization remains largely underexplored.

\subsection{Subject-driven Image Generation}
Given one or more reference images of a subject, subject-driven image generation aims to synthesize images of the same specific subject in diverse contexts and scenarios described by a text prompt. This task has garnered significant attention and has been extensively studied in recent years. Existing approaches can be broadly categorized into two groups: finetuning-based methods \cite{Ruiz_2023_CVPR, gal2022textual, kumari2022customdiffusion, chen2023disenbooth, Chae_2025_CVPR, Nguyen_2025_ICCV} and training-based methods \cite{ye2023ip-adapter, Wei2023ELITEEV, Zhang_2024_CVPR, he2025disenvisioner, li2023blip}.

\myheading{Finetuning-based methods.} These works require test-time finetuning of a pretrained generative model using one or multiple reference images of the target subject. DreamBooth \cite{Ruiz_2023_CVPR} finetunes the entire denoising network with a prior preservation loss to avoid overfitting, while Textual Inversion \cite{gal2022textual} optimizes only a single token to represent the subject. Subsequent methods improve efficiency by finetuning only selected parameters \cite{kumari2022customdiffusion, hu2022lora}, disentangling irrelevant attributes using contrastive objectives \cite{chen2023disenbooth}, or aligning attention to better leverage prior knowledge \cite{Chae_2025_CVPR}. Despite these refinements, test-time finetuning typically takes minutes per subject, making these methods slow and impractical for real-time personalization.

\myheading{Training-based methods.} In contrast to finetuning-based methods, training-based approaches \cite{li2023blip, ye2023ip-adapter,Zhang_2024_CVPR,Wei2023ELITEEV, he2025disenvisioner} enable zero-shot personalization without test-time finetuning by learning a subject encoder that injects identity features into the generative model. BLIP-Diffusion \cite{li2023blip} uses a Q-Former to extract text-aligned visual features from BLIP-2, while IP-Adapter \cite{ye2023ip-adapter} leverages CLIP features through decoupled cross-attention. ELITE \cite{Wei2023ELITEEV} maps visual concepts into textual embeddings, and SSR-Encoder \cite{Zhang_2024_CVPR} learns multi-scale subject embeddings integrated via trainable cross-attention layers. Although avoiding costly test-time finetuning, these models require large-scale personalized datasets, which are difficult to collect, and still rely on slow multi-step diffusion sampling. In this work, we propose a new training-based approach that relies only on readily available image–caption pairs, making data preparation straightforward. Moreover, we further enhance computational efficiency by reducing generation to a single step, achieving real-time personalized image synthesis.

\begin{figure}[t]
    \centering
    \includegraphics[width=\columnwidth]{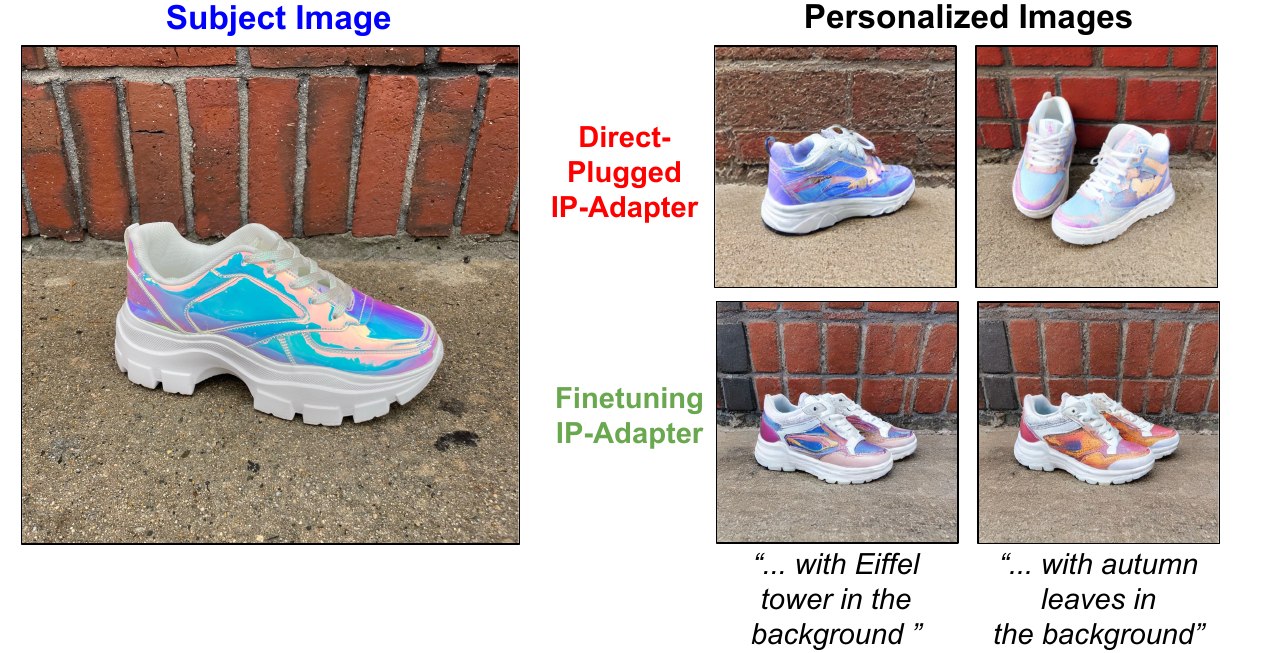}
    \caption{Both the direct-plugged (top) and finetuned (bottom) IP-Adapter variants fail to preserve subject identity and maintain prompt alignment when used with a one-step diffusion model.}
    \label{fig:baseline}
    \vspace{-10pt}
\end{figure}

%% file: sec/3_preliminaries.tex
\section{Preliminaries}
\label{sec:preliminaries}

\subsection{Text-to-image Diffusion Model}
\label{sec:preliminaries_diff}
Text-to-image diffusion models \cite{Rombach_2022_CVPR, 10.1145/3680528.3687625, 10.5555/3600270.3602913, yin2024improved} learn to produce high-quality image outputs that align with conditional text prompt inputs. While early works \cite{10.5555/3495724.3496298, song2022denoisingdiffusionimplicitmodels} perform the diffusion process in image space, most later works perform it in a compact latent space defined by a pretrained VAE for efficient computation. 
In this work, we focus on latent diffusion models.

\minisection{One-step diffusion model}.  Given text prompt $\y_\text{txt}$ and a random Gaussian noise $\epsilon \sim \mathcal{N}(0,I)$, a one-step diffusion model directly predicts a clean image latent $\hat{\z} = \mG_\theta(\epsilon, \y_\text{txt})$ without performing iterative denoising process. 

\subsection{Attention Layers in Diffusion Model}
\label{sec:preliminaries_attention}
Attention layers play a crucial role in incorporating conditional guidance into the outputs generated by diffusion models. Formally, let $\h^{l-1}$ denotes the hidden features from the previous $(l-1)^{\text{th}}$ layer of the denoising network. The hidden features at the $l^{\text{th}}$ layer are then computed via attention as follows:
\begin{align}
\h^{l} = \operatorname{Attn}(\Q^l, \K^l, \V^l)= \operatorname{SoftMax}\left(\frac{\Q^l {\K^l}^{\top}}{\sqrt{d^l}}\right) \V^l
\label{eq:ori_attn},\\
\Q^l = \W^l_\Q \h^{l-1}, \quad \K^l = \W_\K^l \e, \quad \V^l = \W_\V^l \e,
\end{align}
where $d^l$ is the feature dimension at layer $l$. Here, the query features $\Q^l$ are obtained by projecting $\h^{l-1}$ through a weight matrix $\W^l_\Q$, while the key and value pairs, i.e., $\K^l$ and $\V^l$, are derived by projecting a conditional embedding $\e$ through weight matrices $\W^l_\K$ and $\W^l_\V$. Both self-attention and cross-attention layers are implemented using \cref{eq:ori_attn} but with different conditional embeddings. Specifically, self-attention uses $\e = \h^{l-1}$ to capture spatial self-similarity among local feature regions within an image patch, while cross-attention layers incorporate textual embedding $\e = \c_\y$ to condition visual features on the input text prompt $\y_\text{txt}$.
\subsection{Image-Prompt Adapter (IP-Adapter)}
\label{sec:preliminaries_ipa}
A reference image $\x_\text{ref}$ can be incorporated into a pre-trained diffusion model as additional visual guidance by jointly conditioning text and image features via a cross-attention mechanism. To achieve this, IP-Adapter \cite{ye2023ip-adapter} adds extra attention computations inside each cross-attention layer. Given the reference image feature $\c_\x$ (extracted by the CLIP image encoder) and the text feature $\c_\y$ (extracted by the CLIP text encoder), the hidden state $\h^l$ of the cross-attention, as defined in \cref{eq:ori_attn}, can be reformulated as:
\begin{align}
\h^{l} &= \operatorname{Attn}(\Q^l, \K^l_\y, \V^l_\y) + \s_{\x}\operatorname{Attn}(\Q^l, \K^l_\x, \V^l_\x)
\label{eq:ipa_attn},
\end{align}
where $\operatorname{Attn}(.)$ denotes the attention operation described in \cref{sec:preliminaries_attention}. A scaling factor $\s_{\x}$ is introduced to control the influence of $\c_\x$ on the generated output. The key and value pairs for the text features $\c_\y$ are defined as $\K^l_\y = \W^l_{\K,\y} \c_\y$ and $\V^l_\y = \W^l_{\V, \y} \c_\y$, while the key and value pairs for the image features $\c_\x$ are $\K^l_\x = \W^l_{\K,\x} \c_\x$ and $\V^l_\x = \W^l_{\V,\x} \c_\x$.

\subsection{Baseline: Identity Injection with IP-Adapter}
\label{sec:baseline}
A straightforward approach to incorporate subject identity into a one-step diffusion model is through IP-Adapter. However, IP-Adapter is trained to capture global image features and lacks mechanisms to emphasize subject-specific details. To demonstrate this limitation, we conduct two experiments on a one-step diffusion model: (1) directly plugging the IP-Adapter (trained on a multi-step denoiser) and (2) fine-tuning it on a one-step network with an additional perceptual similarity loss (e.g., DIST \cite{9298952}). As shown in \cref{fig:baseline}, both approaches produce suboptimal results—capturing irrelevant background content instead of the subject and failing to align well with the text prompt. We attribute this to two main issues: (i) IP-Adapter’s multi-step training objective does not transfer effectively to the one-step setting, where guidance occurs in a single forward pass, and (ii) fusing image and text features within the same cross-attention layers causes modality competition, degrading both semantic alignment and identity preservation. To overcome these limitations, we introduce a novel dual-branch identity injection architecture combined with mask-guided rescaling, described in the following sections.

%% file: sec/4_methods.tex
\section{Our Approach: Dual-Branch Identity Injection via Attention Features}
\label{sec:dual_branch}
\minisection{Problem Statement.} Our goal is to achieve one-step subject-driven image generalization with diffusion model. To achieve this, we seek an effective injection mechanism to incorporate subject identity into the one-step model. Formally, given the reference subject image $\x_\text{ref}$, text prompt $\y_\text{txt}$, and random Gaussian noise $\epsilon \sim \mathcal{N}(0,I)$, we adapt the one-step text-to-image generator $\mG_\theta(\cdot)$ to incorporate subject image, which generates personalized image latent as:
\begin{equation}
\label{eq:pred_latent}
\hat{\z} = \mG_\theta(\epsilon, \y_\text{txt}, \x_\text{ref}).  
\end{equation}

\myheading{Overview architecture}: We propose an architecture consisting of a One-step Diffusion Model to address the limitations of the stand-alone IP-Adapter, a Reference Network (sharing the same architecture with the one-step model but initialized with the weights of multi-step model), and an IP-Adapter. This dual-branch design effectively incorporates essential identity features for one-step personalization. The key idea is to decouple identity incorporation into two branches: the Reference Network captures \textbf{fine-grained local identity details} injected through self-attention, while the IP-Adapter encodes \textbf{coarse semantic identity features} injected through cross-attention (see \cref{fig:main_diagram}).

\subsection{Identity Injection at Self-Attention}
\label{subsec:self_features}
Self-attention layers in diffusion models are known to capture local self-similarity and preserve high-frequency details \cite{Tumanyan_2023_CVPR, nguyen2023dataset}.
Building on this insight, our first identity injection branch leverages self-attention features for fine-grained identity control. Specifically, we extract key–value (KV) pairs from the self-attention layers of a frozen reference network that processes the reference subject image $\x_\text{ref}$ and inject them into the corresponding layers of the one-step target model to ensure feature alignment.
The reference network shares the same architecture as the one-step model and corresponds to the multi-step teacher used during its training.
To obtain these features, we first extract the subject foreground from $\x_\text{ref}$, encode it with a VAE to obtain $\z_\text{ref}$, and perturb it with Gaussian noise at timestep $t=1$ to preserve structural and identity information. A single forward pass through the frozen reference denoiser then produces the KV features used for injection.

As a result, a set of reference KV self-attention features from different layers $l$, denoted as $(\K^l_\text{ref}, \V^l_\text{ref})$, is fused with the self-attention KV features of the one-step denoising network $\mG_\theta(\cdot)$, denoted as $(\K^l_\text{one}, \V^l_\text{one})$, to guide the generation process.
Specifically, we construct an augmented triplet $(\Q^l_\text{one}, \K^l_\text{one} \oplus \K^l_\text{ref}, \V^l_\text{one} \oplus \V^l_\text{ref})$, where $\oplus$ indicates concatenation along the token dimension, to replace the original triplet $(\Q^l_\text{one}, \K^l_\text{one}, \V^l_\text{one})$ and recompute the hidden states according to the attention formulation in \cref{sec:preliminaries_attention}.
This injection mechanism allows the one-step model to propagate detailed identity cues through local self-similarity, thereby reinforcing fine-grained identity consistency and preserving the subject’s texture fidelity.


\subsection{Identity Injection at Cross-Attention} 
\label{subsec:cross_features}
The second branch design focuses mainly on more generic features, such as pose and shape of the subject identity. We adopt the IP-Adapter injection mechanism as discussed in \cref{sec:preliminaries_ipa}, which combines textual and visual embeddings to provide semantic and contextual guidance for the overall generation process. 
In sum, the self-attention branch refines detailed visual structures, while the cross-attention branch ensures that the generated output remains faithful to both the textual description and the subject’s identity. 

\begin{table*}[ht]
\centering
\small
\setlength{\tabcolsep}{8pt}
\caption{Quantitative comparison with several multi-step personalized image generation approaches on DreamBench.}
\begin{tabular}{llccccr} 
\toprule
\multirow{2}{*}{\textbf{Type}} & \multirow{2}{*}{\textbf{Method}} & \multicolumn{3}{c}{\textbf{Subject Preservation}} & \multicolumn{1}{c}{\textbf{Text Alignment}} & \multirow{2}{*}{\textbf{Runtime} $\downarrow$} \\ 
\cmidrule(lr){3-5}
\cmidrule(lr){6-6} 
& & CLIP-I $\uparrow$ & DINO $\uparrow$ & Nexus $\uparrow$ & CLIP-T $\uparrow$ & (seconds) \\ 
\midrule
\multirow{5.6}{*}{\begin{tabular}[c]{@{}l@{}}\textbf{Multi-step}\\\textbf{(50 steps)}\end{tabular}} 
& DreamBooth \cite{Ruiz_2023_CVPR} & 0.800  & 0.632 & 0.692 & \underline{0.306} & 274.00s \\
& IP-Adapter \cite{ye2023ip-adapter} & 0.848 & 0.728 & 0.760 & 0.284 & 3.88s  \\ 
& ELITE \cite{Wei2023ELITEEV} & 0.794 & 0.620 & 0.663 & 0.283 & 3.91s \\
& SSR-Encoder \cite{Zhang_2024_CVPR} & 0.800 & 0.637 & 0.689 & 0.304 & 4.28s \\ 
& BLIP-Diffusion \cite{li2023blip} & 0.810 & 0.618 & 0.702 & 0.284 & 3.77s \\
& DisEnvisioner \cite{he2025disenvisioner} & \underline{0.858} & \underline{0.752} & \underline{0.765} & \textbf{0.309} & 4.27s \\
\midrule
\multirow{1}{*}{\begin{tabular}[c]{@{}l@{}}\textbf{One-step}\\\end{tabular}}
& SwiftPie (proposed) & \textbf{0.862} & \textbf{0.777} & \textbf{0.778} & \underline{0.306} & \textbf{0.17s} \\
\bottomrule
\end{tabular}
\label{tab:main_quant}
\vspace{-10pt}
\end{table*}

\subsection{Training Strategy}
\minisection{Training objectives.} 
We train our dual-branch injection framework using weak reconstruction objectives, including a \textbf{perceptual loss in image space} and an \textbf{adversarial loss in latent space}.
Importantly, we do not employ a strong L2 reconstruction loss, as our goal is image generation rather than exact reconstruction.
This design allows different random noise samples $\epsilon$ to produce images that are perceptually aligned, preserving the core subject identity of the reference subject. We leverage these losses for LoRA training of the one-step generator $\mG$ and the linear projection layer of the IP-Adapter. Note that we only need pairs of images and captions to train our network.

For the \textbf{perceptual loss in image space}, we predict the image latent $\z_\text{rec}$ (from \cref{eq:pred_latent}) using the one-step network and decode it through the VAE decoder to obtain the reconstructed image $\x_\text{rec}$.
To align the reconstructed image with the reference, we apply the \textbf{DIST} \cite{9298952} and \textbf{SSIM} \cite{1284395} loss functions, which encourage semantic consistency and structural fidelity, respectively.
The overall perceptual loss is formulated as:
\begin{equation}
    \mL_{\text{perceptual}} = \lambda_1\mL_{\text{DIST}}(\x_\text{ref}, \x_\text{rec}) + \lambda_2\mL_{\text{SSIM}}(\x_\text{ref}, \x_\text{rec}),
\label{eq:perceptual}
\end{equation}
where $\lambda_1$ and $\lambda_2$ are hyper-parameters controlling the contribution of each term. 

Beside perceptual loss, we adopt an \textbf{adversarial objective in latent space} to improve image fidelity further. Following \cite{10.1145/3680528.3687625}, we use the diffusion encoder, along with a classification head adding on top, as a discriminator to distinguish perturbed real and fake samples. The discriminator encoder also uses the reference image as condition using decoupled cross-attention. Given the reference latent $\z_{\text{ref}}$ obtained from the VAE encoder and the reconstructed latent $\z_{\text{rec}}$ predicted as in \cref{eq:pred_latent}, we perturb both via a forward diffusion process $F$ (i.e., noise injection) at a random timestep $t \sim [0, T]$, yielding $\z^t_{\text{ref}}$ and $\z^t_{\text{rec}}$. These noisy latents are then passed through the conditional diffusion discriminator $\mD$, which is trained to distinguish fake noisy samples $\z^t_{\text{rec}}$ from real noisy samples $\z^t_{\text{ref}}$. Specifically, we train the classification head along with diffusion encoder and the one-step generator $\mG$ by minimizing the following GAN objective:
\vspace{-15pt}
\begin{align}
    &\z^t_{\text{ref}} = F(\z_{\text{ref}}, t), \quad \z^t_{\text{rec}} = F(\z_\text{rec}^t, t) \nonumber \\
    &\mL^{\mD}_{\text{adv}} = - \mathbb{E}_{t,\, \z^t_{\text{ref}},\, \z^t_{\text{rec}}} 
    \Big[
        \log \mD(\z^t_{\text{ref}})
        + \log \big(1 - \mD(\z^t_{\text{rec}})\big)
    \Big] \nonumber \\
    &\mL^{\mG}_{\text{adv}} = - \mathbb{E}_{t,\, \z^t_{\text{rec}}} \log \mD(\z^t_{\text{rec}}).
\label{eq:loss_dis}
\end{align}
%
%
The final loss function for training one-step generator $\mG$ is:
\begin{equation}
\mL^{\mG}_{\text{total}} = \mL_\text{perceptual} + \lambda_3\mL^{\mG}_{\text{adv}},
\label{eq:loss_gen}
\end{equation}
where $\lambda_3$ control influence of adversarial loss.

\myheading{Curriculum training stage for subject variation.} 
The second branch captures general attributes like pose, shape, and viewpoint, but excessive conditioning can cause overfitting and limit variation. To mitigate this, we introduce a second training phase with curriculum learning, following the “easy-to-hard” principle \cite{bengio2009curriculum, wang2021survey,Mao_2025}. In the first stage, the dual-branch framework focuses on identity preservation; once identity features are well learned, the second stage applies random cropping to the input image of the IP-adapter branch, enhancing generalization across diverse poses and views. At each training iteration $i$, we define a crop scale interval $[r_i, r_{max}]$ and sample the random crop scale as $r_{\text{crop}} \sim U(r_i, r_{max})$, where
\begin{equation}
    r_i = \lambda^{i/I}(r_{max}-r_{min})  + r_{min}.
\label{eq:crop_scale}
\end{equation}
Here, $\lambda^{i/I}$ is a dynamic scale that starts at $\lambda^0 = 1$ and decays exponentially to $\lambda$ by the final iteration $I$. Meanwhile, $r_\text{min}$ and $r_\text{max}$ are the minimum and maximum scales of the crop from the reference image that we will use as input. 


\begin{figure*}[ht]
    \centering
    \includegraphics[width=.85\textwidth]{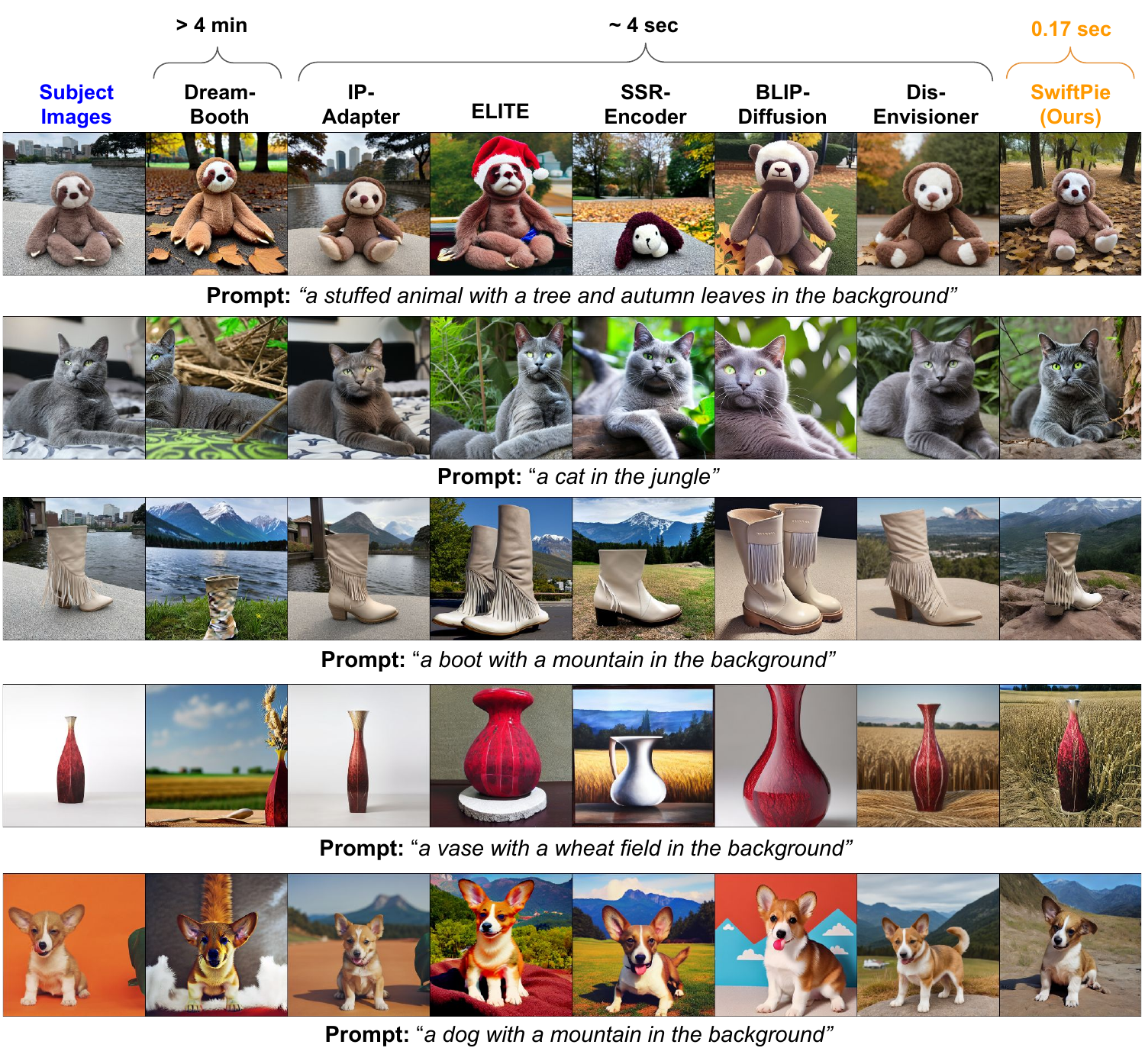}
    \vspace{-5pt}
    \caption{Qualitative comparison with other approaches on DreamBench.}
    \vspace{-10pt}
    \label{fig:main_qual}
\end{figure*}

\subsection{Inference Strategy}

\label{sec:mask_guided}
To further enhance contextualization guided by the text prompt, we introduce a \textbf{mask-guided rescaling technique} at inference time.
Inspired by prior works \cite{nguyen2023dataset, hertz2023prompttoprompt}, we compute a \textbf{subject mask} by leveraging the cross-attention map obtained from \cref{sec:preliminaries_attention}, which captures the spatial correlation between the subject token in the prompt and the generated image (e.g., `dog' in ``A dog sitting on a grass field'').
The mask is then binarized with a threshold of $\tau=0.6$ and resized to match the spatial resolution of the corresponding feature map at the $l^{th}$ layer, yielding $\M^l \in \{0,1\}^{H^l\times W^l}$.
During the forward pass of each cross-attention layer, we reformulate the hidden-state computation to incorporate mask-guided rescaling.
The core idea is to spatially modulate the attention mechanism using the subject mask -- amplifying the influence of subject features within the masked region while allowing background regions to remain guided by the text prompt. 
Formally, instead of the standard cross-attention described in \cref{eq:ipa_attn}, the hidden state is updated as:
{ \small 
\begin{align}
\h^{l} &= \operatorname{Attn}(\Q^l, \K^l_\y, \V^l_\y) +\s_{\x} \M^l \odot \operatorname{Attn}(\Q^l, \K^l_\x, \V^l_\x).
\label{eq:mask_guided_rescale}
\end{align}
}
where $\odot$ indicates element-wise multiplication.


%% file: sec/5_experiments.tex
\section{Experiments}

\label{sec:experiments}

\begin{figure*}[ht]
    \centering
    \includegraphics[width=0.8\textwidth]{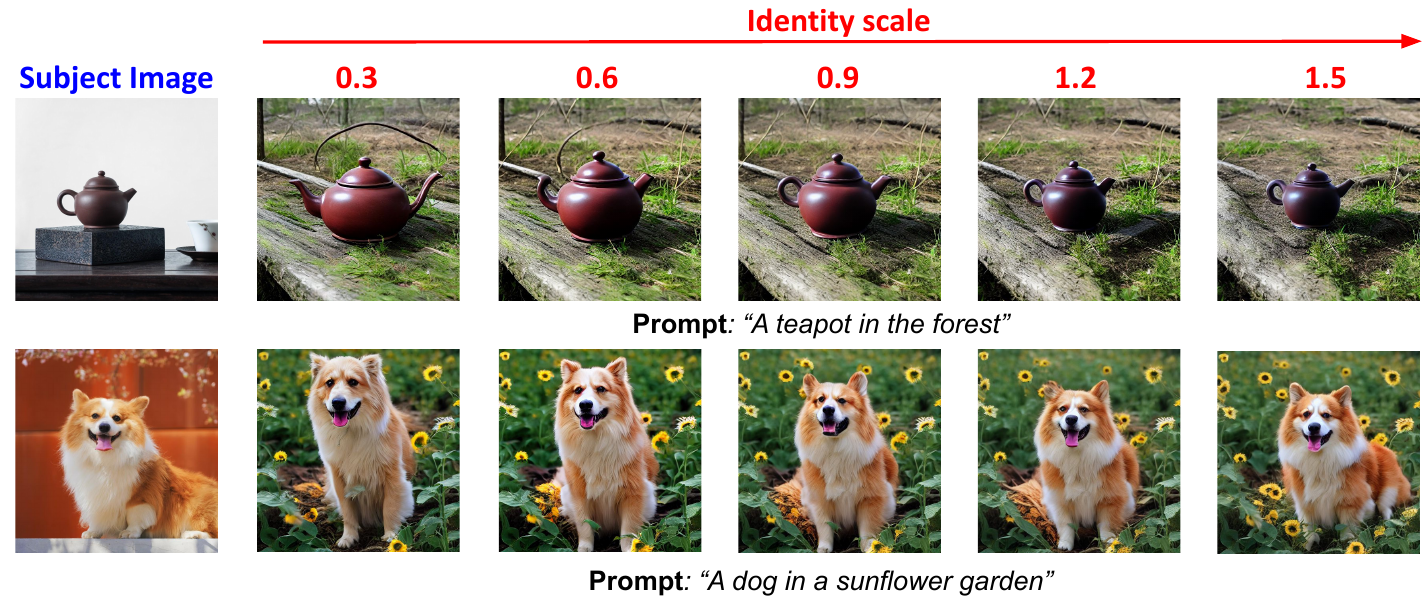}
    \vspace{-5pt}
    \caption{Varying subject identity scale used in mask-guided rescaling}
    \vspace{-10pt}
    \label{fig:ablate_scale}
\end{figure*}

\begin{figure}[t]
    \centering
    \includegraphics[width=\columnwidth]{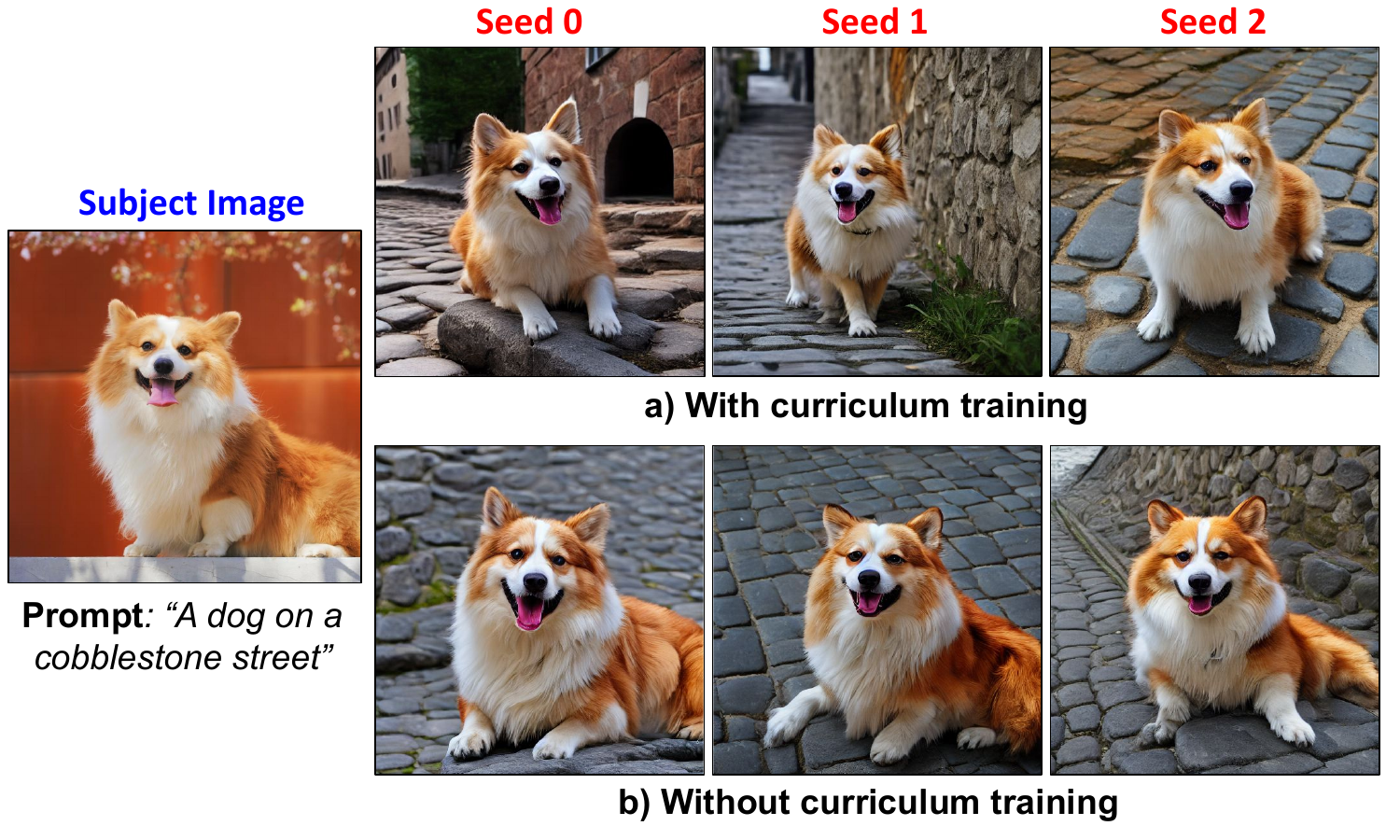}
    \caption{Each column corresponds to a different random seed used for personalized generation. Top: with curriculum training. Bottom: without curriculum training.}
    \label{fig:ablate_curriculum}
\end{figure}


\minisection{Datasets}. We train our framework on the DreamCache dataset \cite{Aiello_2025_CVPR}, which contains 400K (image, caption, mask) triplets generated using SDXL \cite{podell2024sdxl} and Lang-SAM pipeline \cite{langsegmentanything} for subject mask extraction. We follow prior works \cite{Wei2023ELITEEV, Ruiz_2023_CVPR, he2025disenvisioner} to evaluate on DreamBench \cite{Ruiz_2023_CVPR}, covering 30 subjects with 3–5 reference images each and 25 text prompts, resulting in 3950 generated images for comparison. Note that we only use one reference image per text prompt. 

\minisection{Metrics}.
We evaluate performance using two widely adopted metrics for subject-driven image personalization:
(1) Identity preservation, measured by CLIP-I \cite{Radford2021LearningTV}, DINO \cite{Caron_2021_ICCV}, and NexusScore \cite{yuan2025opensvnexus}. Subject regions are first localized using an image–prompt detection model, and visual embeddings are extracted with GME-Qwen2VL \cite{zhang2024gme}; similarity is then computed in the image feature space.
(2) Text–image alignment, measured by CLIP-T \cite{Radford2021LearningTV}.


\minisection{Implementation details.} All experiments are conducted on a single NVIDIA A100 GPU (40GB). We train with batch size of four and gradient accumulation of two using AdamW optimizer \cite{loshchilov2018decoupled} for 100K and 60K iterations in two stages, with learning rates of $10^{-5}$ and $10^{-6}$. The second stage performs light finetuning for better generalization while preserving identity. Curriculum parameters are $r_{\min}=0.1$, $r_{\max}=1$, and $\lambda=0.025$, and perceptual/ generation losses use $\lambda_1=1$, $\lambda_2=0.2$, $\lambda_3=0.1$. We adopt a one-step diffusion model based on \textbf{SD 1.5}, following SBv2 \cite{dao2025swiftbrush}, with the IP-Adapter Plus variant. Trainable parameters include LoRA \cite{hu2022lora} applied to attention layers (to\_q, to\_k, to\_v, to\_out.0) of both one-step model and diffusion discriminator model (rank of 64), 
together with IP-Adapter projection layers and discriminator heads, totaling approximately 120M parameters.

\minisection{Baselines}.
We compare our method against several encoder-based multi-step personalization approaches, including IP-Adapter \cite{ye2023ip-adapter}, DisEnvisioner \cite{he2025disenvisioner}, ELITE \cite{Wei2023ELITEEV}, BLIP-Diffusion \cite{li2023blip}, and SSR-Encoder \cite{Zhang_2024_CVPR}, as well as the widely used fine-tuning–based method DreamBooth \cite{Ruiz_2023_CVPR}. For a fair comparison, all methods use the same Stable Diffusion 1.5 backbone and the default setting of 50 sampling steps.

\subsection{Comparative Results}
\minisection{Qualitative Results.} In \cref{fig:main_qual}, we visualize personalized results of various multi-step approaches, comparing them with \Approach. As shown, \Approach~   consistently achieves strong personalization with high-quality generated results. \Approach~can capture high fidelity subject preservation while aligning well with guided text prompts, shown below each image. Notably, \Approach~achieves this performance with a single forward pass, generating each personalized image in just 0.17 seconds, much faster than training-based multi-step methods, which typically require over 4s, and fine-tuning-based approaches, which often take over 4 mins per image.

\minisection{Quantitative Results.} As shown in \cref{tab:main_quant}, \Approach~is the first to achieve one-step personalization. In addition to its remarkable speed advantage, \Approach~delivers strong quantitative performance, outperforming all multi-step approaches by a large margin in subject preservation and achieving the second-best CLIP-T score for text alignment.

%% file: sec/6_ablate.tex
\subsection{Ablation studies}
\label{sec:ablation_studies}
Next, we conduct several ablation studies to analyze the effect of different components used in \Approach. 

\minisection{Impact of dual-branch injection design} is reported in \cref{tab:ablate_quan}. 
Our proposed dual-branch injection achieves the best identity preservation score. When incorporating the 
\textbf{Mask-guided Rescaling} strategy at inference time, the text alignment score (CLIP-T) substantially improves, demonstrating its effectiveness in contextualization enhancement. While the identity score decreases slightly, SwiftPie still outperforms remaining settings and other multi-step methods in subject preservation. Additionally, \cref{fig:ablate_scale} illustrates the effect of the identity scale in \cref{eq:mask_guided_rescale}, with larger scales yielding stronger identity preservation.


\begin{table}[t]
\centering
\small
\setlength{\tabcolsep}{5pt}
\caption{Ablation studies on our components}
\begin{tabular}{lcccc} 
\toprule
\textbf{Setting} & CLIP-I$\uparrow$ & DINO$\uparrow$ & CLIP-T$\uparrow$\\ 
\midrule
Direct plug IP-Adapter & 0.840  & 0.723 & 0.270 \\
Finetune IP-Adapter  & 0.850  & 0.736 & 0.265 \\ 
Single Ref-KV Unet  & 0.843  & 0.735 & 0.286 \\
\midrule
Proposed   & \textbf{0.862}  & \textbf{0.777} & \textbf{0.306}\\ 
\bottomrule
\end{tabular}
\label{tab:ablate_quan}
\vspace{-10pt}
\end{table}


\minisection{Curriculum training improves subject's generalization}. As shown in \cref{fig:ablate_curriculum}, we visualize personalized results generated with different random seeds, comparing models trained with (top row) and without (bottom row) curriculum training. Incorporating curriculum training in our dual-branch identity injection framework enhances the model’s ability to vary the subject’s pose, position, and viewpoint. Without curriculum training, results show limited diversity, with the model struggling to generalize subject variations—crucial for applications requiring flexible and diverse outputs.

\begin{figure}[t]
    \centering
    \includegraphics[width=.95\columnwidth]{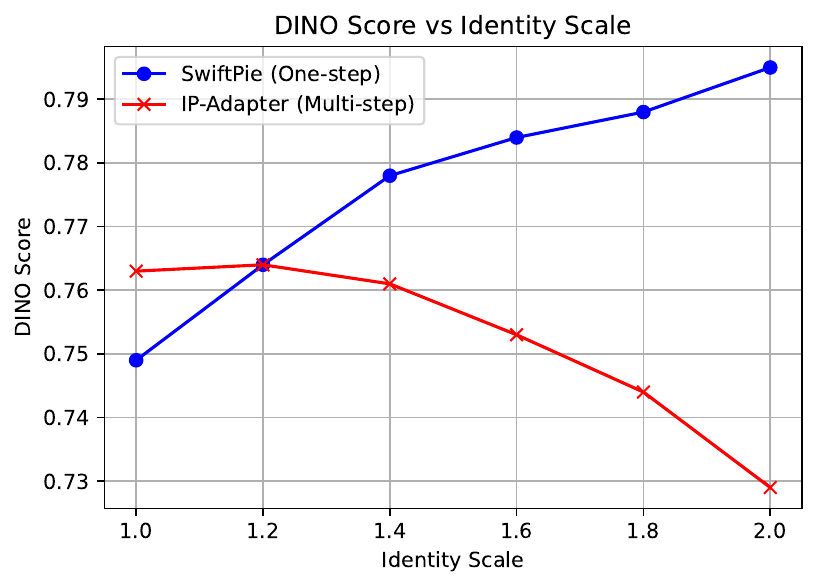}
    \caption{Impact of identity scale on identity preservation.}
    \label{fig:add_ablate_scale}
\end{figure}

\minisection{Effect of identity scale.}
The identity scale in \cref{eq:mask_guided_rescale} is fixed at $\s_{\x}=1.6$ for all experiments. To examine its effect, we compare it with the identity scale used in IP-Adapter \cite{ye2023ip-adapter} by evaluating identity preservation using the DINO score across different scale values. As shown in \cref{fig:add_ablate_scale}, SwiftPie attains lower identity scores than multi-step IP-Adapter at smaller scales (e.g., 1.0), but its identity preservation improves markedly as the scale increases, while IP-Adapter’s performance gradually degrades. This gain is due to our mask-guided rescaling, which selectively injects identity features into the foreground subject, whereas IP-Adapter scales global image features and may amplify background information at larger scales, reducing subject fidelity.

\label{sec:experiments}

%% file: sec/7_conclusion.tex
\section{Conclusion}
\label{sec:conclusion}
In this paper, we introduce \Approach, a one-step diffusion framework that can achieve lightning-fast speed in subject-driven image personalization via \textbf{dual-branch identity injection design} and \textbf{mask-guided rescaling}. While being significantly fast compared to other multi-step counterparts, \Approach~delivers a high-quality result in terms of both subject's identity preservation and text prompt alignment, demonstrated through extensive experimental results. For future work, our aim is to further enhance the quality of the generated results by extending our framework to more powerful, state-of-the-art diffusion backbone such as FLUX\cite{flux2024} or Z-image\cite{team2025zimage}, as well as supporting more complex personalize scenarios such as multiple subject personalization.

%% file: sec/X_suppl.tex
\clearpage
\setcounter{page}{1}
\maketitlesupplementary

\begin{figure*}[t]
    \centering
    \includegraphics[width=\textwidth]{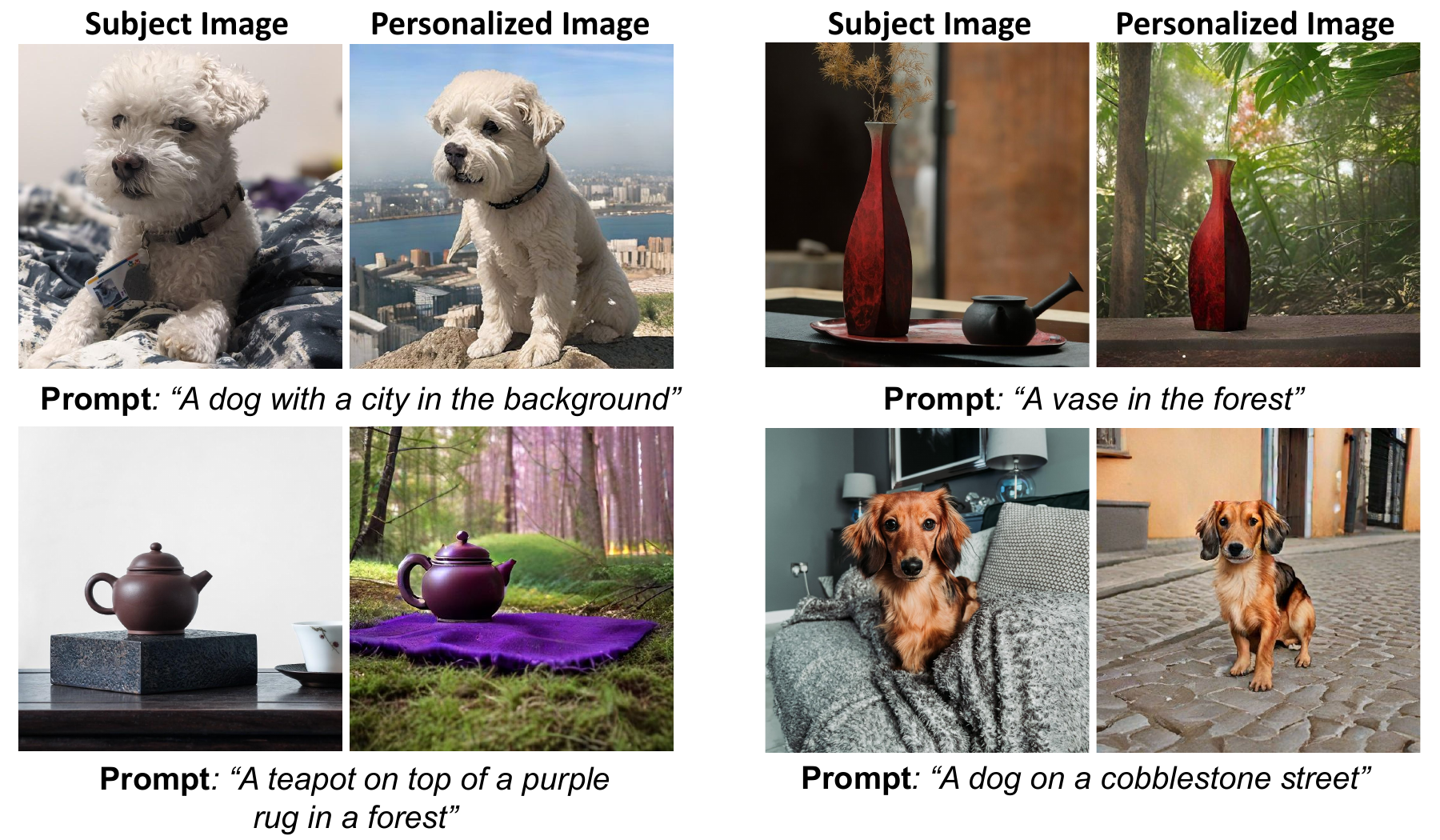}
    \caption{Qualitative results of SwiftPie with DMDv2 one-step model (SDXL backbone) on DreamBench.}
    \label{fig:add_sdxl_qual}
\end{figure*}

In this supplementary material, we present additional quantitative and qualitative results in \cref{sec:add_result}. Furthermore, we provide extended ablation studies in \cref{sec:add_ablate} to examine the contribution of each component in SwiftPie. A user study is also included in \cref{sec:user_study} to assess user's preference of our personalization approach. Finally, we discuss the societal implications of our personalization tool in \cref{sec:soc_impact}.

\section{Additional Results}
\label{sec:add_result}
We report additional quantitative results on DreamBench++ \cite{peng2024dreambench} in \cref{tab:add_quan}. DreamBench++ extends the original DreamBench benchmark \cite{Ruiz_2023_CVPR} by providing more reference images and personalized prompts (9 per subject) generated by GPT-4o. The reference images cover several sub-categories, including objects, animals, humans, and styles. Since our work focuses on subject-driven personalization, we exclude style references from the evaluation. Note that each subject is represented by a single reference image. In the object sub-category, SwiftPie outperforms other multi-step approaches in both subject preservation and text alignment. For animal and human sub-categories, SwiftPie delievers competitive performance relative to multi-step methods. We believe these results could be further improved by expanding SwiftPie's training data to include more examples from these sub-categories.
\begin{table*}[h!]
\centering
\footnotesize
\setlength{\tabcolsep}{8pt}
\caption{Quantitative comparison with several multi-step personalization approaches on DreamBench++. Each color box \colorbox{BlueGreen}{object}/\colorbox{BurntOrange}{animal}/\colorbox{Thistle}{human}indicates benchmark results on different sub-categories.}
\begin{tabular}{llcccc} 
\toprule
\multirow{2}{*}{\textbf{Type}} & \multirow{2}{*}{\textbf{Method}} & \multicolumn{3}{c}{\textbf{Subject Preservation}} & \multicolumn{1}{c}{\textbf{Text Alignment}} \\ 
\cmidrule(lr){3-5}
\cmidrule(lr){6-6} 
& & CLIP-I $\uparrow$ & DINO $\uparrow$ & Nexus $\uparrow$ & CLIP-T $\uparrow$ \\
\midrule
\multirow{5.6}{*}{\begin{tabular}[c]{@{}l@{}}\textbf{Multi-step}\\\textbf{(50 steps)}\end{tabular}} 
& DreamBooth \cite{Ruiz_2023_CVPR} & \colorbox{BlueGreen}{0.586}/\colorbox{BurntOrange}{0.587}/\colorbox{Thistle}{0.488} & \colorbox{BlueGreen}{0.201}/\colorbox{BurntOrange}{0.156}/\colorbox{Thistle}{0.204} & \colorbox{BlueGreen}{0.293}/\colorbox{BurntOrange}{0.302}/\colorbox{Thistle}{0.212} & \colorbox{BlueGreen}{0.287}/\colorbox{BurntOrange}{0.275}/\colorbox{Thistle}{\textbf{0.320}} \\

& IP-Adapter \cite{ye2023ip-adapter} & \colorbox{BlueGreen}{\underline{0.853}}/\colorbox{BurntOrange}{0.889}/\colorbox{Thistle}{\textbf{0.787}} & \colorbox{BlueGreen}{\underline{0.743}}/\colorbox{BurntOrange}{\underline{0.790}}/\colorbox{Thistle}{\underline{0.584}} & \colorbox{BlueGreen}{\underline{0.771}}/\colorbox{BurntOrange}{\textbf{0.819}}/\colorbox{Thistle}{\textbf{0.622}} & \colorbox{BlueGreen}{0.292}/\colorbox{BurntOrange}{0.294}/\colorbox{Thistle}{0.235} \\

& ELITE \cite{Wei2023ELITEEV} & \colorbox{BlueGreen}{0.788}/\colorbox{BurntOrange}{0.849}/\colorbox{Thistle}{0.705} & \colorbox{BlueGreen}{0.578}/\colorbox{BurntOrange}{0.693}/\colorbox{Thistle}{0.489} & \colorbox{BlueGreen}{0.603}/\colorbox{BurntOrange}{0.704}/\colorbox{Thistle}{0.492} & \colorbox{BlueGreen}{0.310}/\colorbox{BurntOrange}{0.311}/\colorbox{Thistle}{0.270} \\

& SSR-Encoder \cite{Zhang_2024_CVPR} & \colorbox{BlueGreen}{0.814}/\colorbox{BurntOrange}{0.851}/\colorbox{Thistle}{0.709} & \colorbox{BlueGreen}{0.628}/\colorbox{BurntOrange}{0.704}/\colorbox{Thistle}{0.461} & \colorbox{BlueGreen}{0.689}/\colorbox{BurntOrange}{0.726}/\colorbox{Thistle}{0.479} & \colorbox{BlueGreen}{0.314}/\colorbox{BurntOrange}{0.316}/\colorbox{Thistle}{0.271} \\

& BLIP-Diffusion \cite{li2023blip} & \colorbox{BlueGreen}{0.828}/\colorbox{BurntOrange}{0.867}/\colorbox{Thistle}{0.713} & \colorbox{BlueGreen}{0.639}/\colorbox{BurntOrange}{0.707}/\colorbox{Thistle}{0.511} & \colorbox{BlueGreen}{0.721}/\colorbox{BurntOrange}{0.759}/\colorbox{Thistle}{0.533} & \colorbox{BlueGreen}{0.286}/\colorbox{BurntOrange}{0.306}/\colorbox{Thistle}{0.237} \\

& DisEnvisioner \cite{he2025disenvisioner} & \colorbox{BlueGreen}{0.851}/\colorbox{BurntOrange}{\textbf{0.892}}/\colorbox{Thistle}{\underline{0.783}} & \colorbox{BlueGreen}{0.725}/\colorbox{BurntOrange}{0.786}/\colorbox{Thistle}{\textbf{0.589}} & \colorbox{BlueGreen}{0.767}/\colorbox{BurntOrange}{\underline{0.808}}/\colorbox{Thistle}{\underline{0.605}} & \colorbox{BlueGreen}{\underline{0.315}}/\colorbox{BurntOrange}{\underline{0.320}}/\colorbox{Thistle}{0.258} \\

\midrule
\multirow{1}{*}{\begin{tabular}[c]{@{}l@{}}\textbf{One-step}\\\end{tabular}}
& SwiftPie (Ours) & \colorbox{BlueGreen}{\textbf{0.857}}/\colorbox{BurntOrange}{\underline{0.890}}/\colorbox{Thistle}{0.746} & \colorbox{BlueGreen}{\textbf{0.770}}/\colorbox{BurntOrange}{\textbf{0.815}}/\colorbox{Thistle}{0.540} & \colorbox{BlueGreen}{\textbf{0.783}}/\colorbox{BurntOrange}{0.796}/\colorbox{Thistle}{0.534} & \colorbox{BlueGreen}{\textbf{0.316}}/\colorbox{BurntOrange}{\textbf{0.321}}/\colorbox{Thistle}{\underline{0.272}} \\
\bottomrule
\end{tabular}
\vspace{-5pt}
\label{tab:add_quan}
\vspace{-10pt}
\end{table*}

\begin{figure*}[h]
    \centering
    \includegraphics[width=\textwidth]{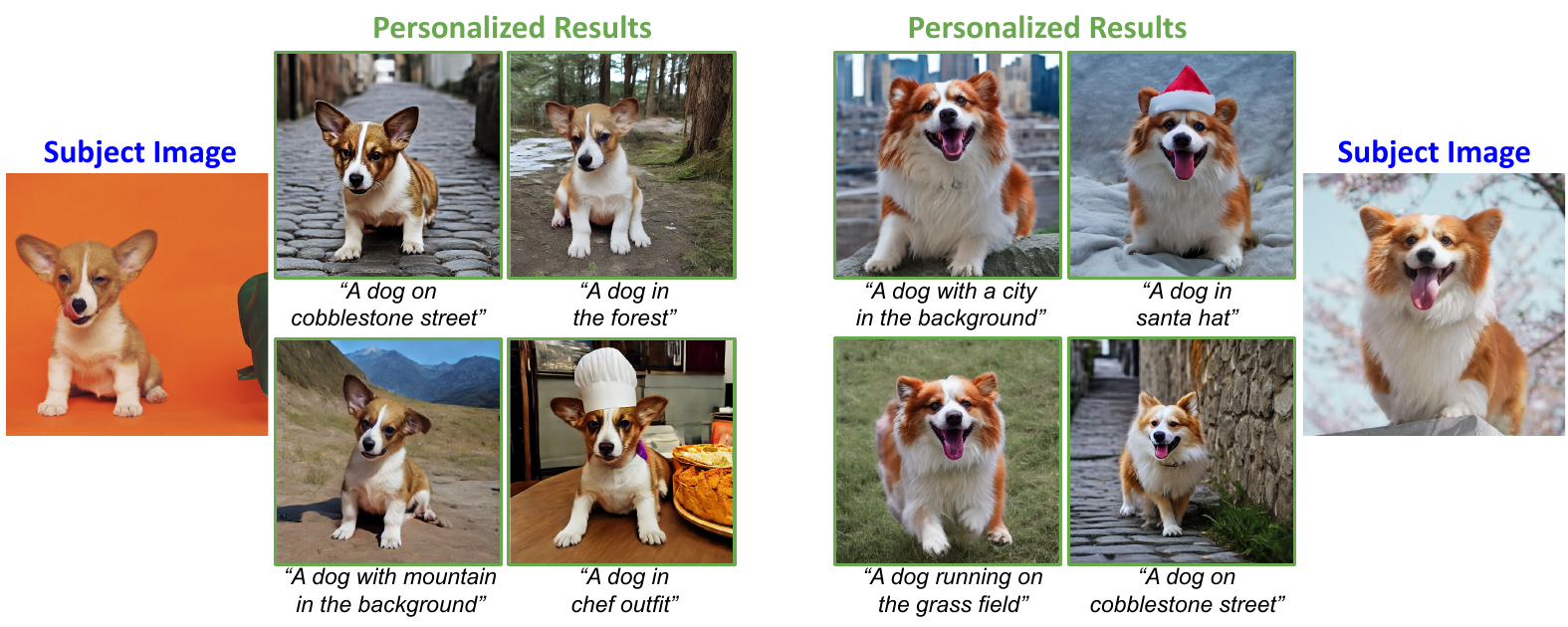}
    \caption{Additional personalization results on DreamBench.}
    \label{fig:add_qual_1}
\end{figure*}

\begin{figure*}[h]
    \centering
    \includegraphics[width=\textwidth]{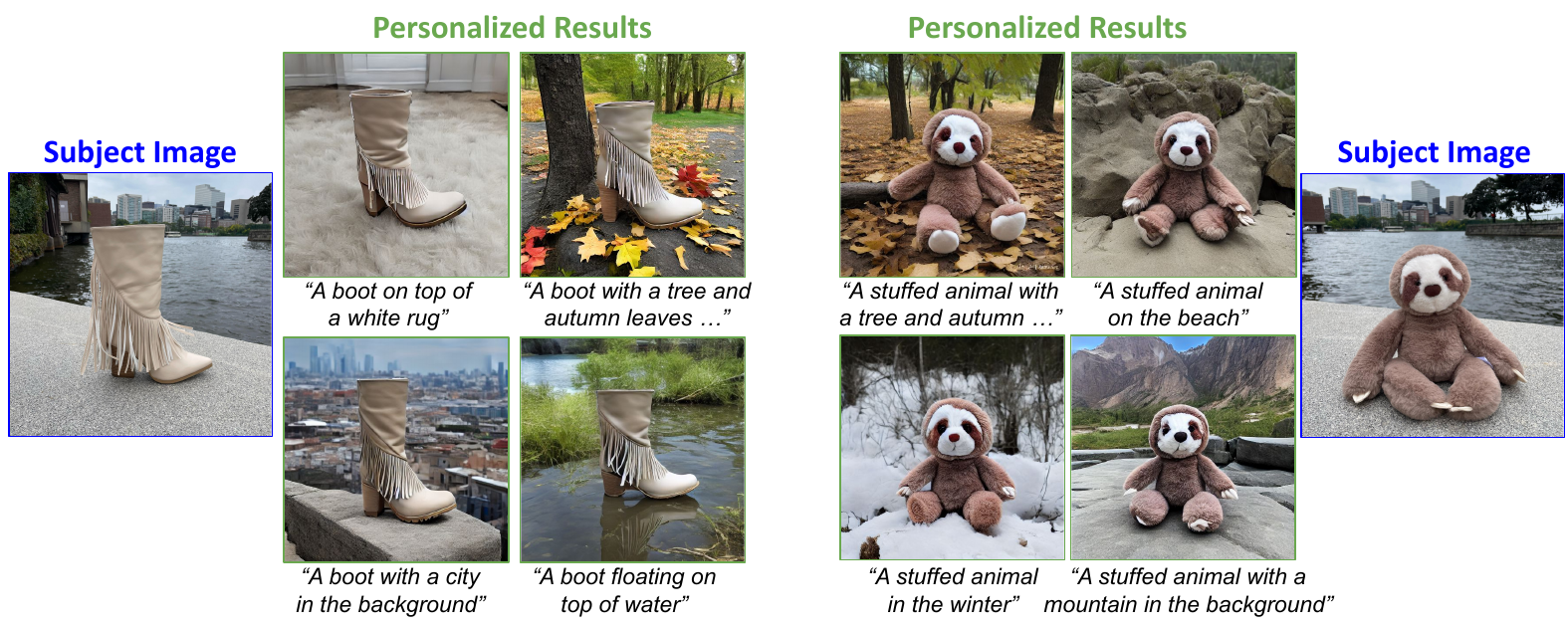}
    \caption{Additional personalization results on DreamBench.}
    \label{fig:add_qual_2}
\end{figure*}

\begin{figure*}[h]
    \centering
    \includegraphics[width=\textwidth]{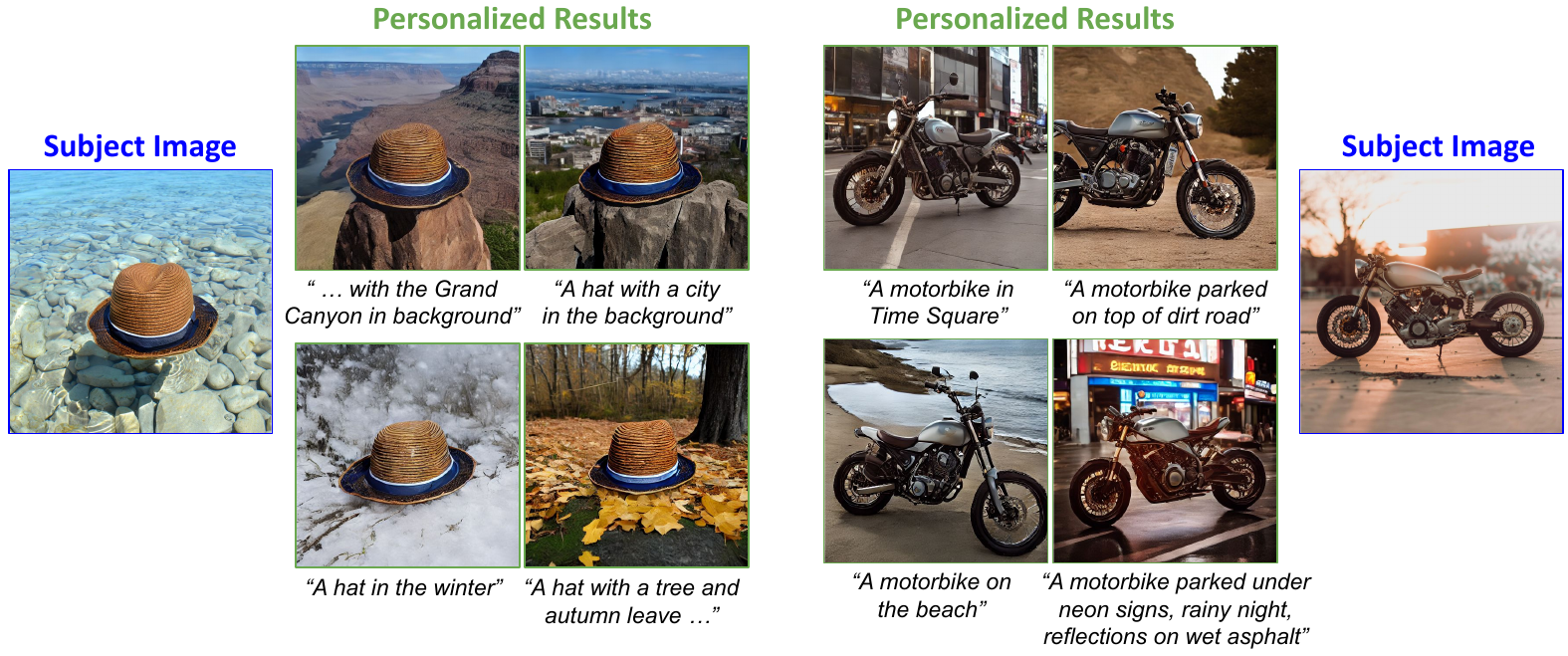}
    \caption{Additional personalization results on DreamBench++.}
    \label{fig:add_qual_3}
\end{figure*}

\begin{figure*}[h]
    \centering
    \includegraphics[width=\textwidth]{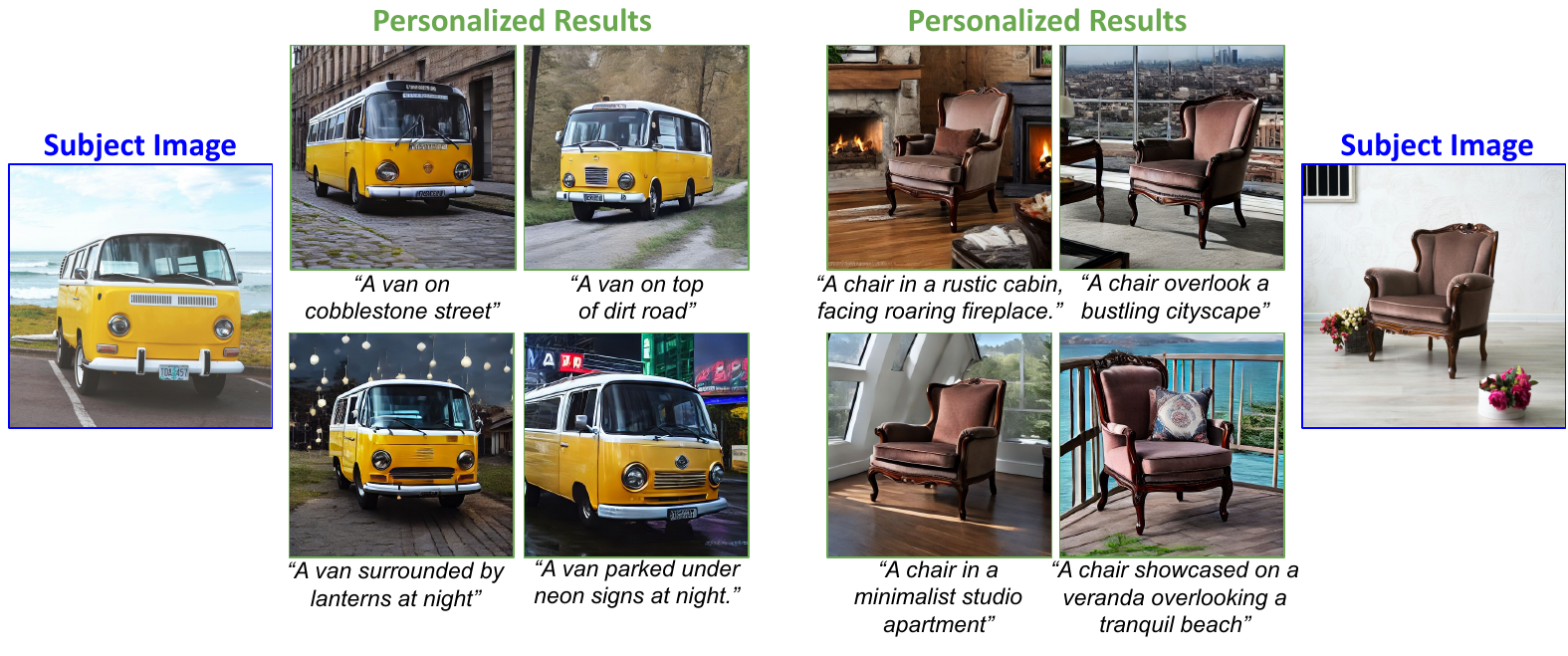}
    \caption{Additional personalization results on DreamBench++.}
    \label{fig:add_qual_4}
\end{figure*}

Additionally, we provide more qualitative results of SwiftPie on both DreamBench \cite{Ruiz_2023_CVPR} (see \cref{fig:add_qual_1}, \cref{fig:add_qual_2}), and DreamBench++ \cite{peng2024dreambench} (see \cref{fig:add_qual_3}, \cref{fig:add_qual_4}). As illustrated, our one-step personalization approach consistently delivers high-fidelity subject preservation and strong text alignment, while maintaining significantly faster inference compared to multi-step methods. The qualitative examples also demonstrate that SwiftPie effectively handles diverse subject categories and complex prompts without sacrificing visual quality.

\section{Additional Ablation Studies}
\label{sec:add_ablate}
\subsection{Effect of losses}
We investigate the impact of different training objective configurations used for SwiftPie. Specifically, we ablate the contributions of both perceptual losses and adversarial loss during training. As shown in \cref{tab:add_ablate_loss}, removing either perceptual loss (DIST and SSIM) or the adversarial loss results in a significant drop in identity preservation, whereas the full objective setting achieves the best identity preservation score.

\begin{table*}[ht]
\centering
\small
\setlength{\tabcolsep}{8pt}
\caption{Ablation studies on different training objectives.}
\begin{tabular}{cccccccc} 
\toprule
\multirow{2}{*}{\textbf{Setting}} & \multirow{2}{*}{$\mathcal{L}_{\text{SSIM}}$} & \multirow{2}{*}{$\mathcal{L}_{\text{DIST}}$} & \multirow{2}{*}{$\mathcal{L}_{\text{adv}}$} & \multicolumn{3}{c}{\textbf{Subject Preservation}} & \multicolumn{1}{c}{\textbf{Text Alignment}} \\ 
\cmidrule(lr){5-7}
\cmidrule(lr){8-8} 
& & & & CLIP-I $\uparrow$ & DINO $\uparrow$ & Nexus $\uparrow$ & CLIP-T $\uparrow$ \\
\midrule
Setting 1 & \cmark & \xmark & \cmark & 0.824 & 0.684 & 0.720 & \textbf{0.313} \\
Setting 2 & \xmark & \cmark & \cmark & 0.842 & 0.729 & 0.744 & 0.310 \\
Setting 3 & \cmark & \cmark & \xmark & 0.841 & 0.736 & 0.748 & 0.311 \\
Setting 4 & \xmark & \xmark & \cmark & 0.824 & 0.683 & 0.719 & 0.310 \\
\textbf{Full Setting (SwiftPie)} & \cmark & \cmark & \cmark & \textbf{0.862} & \textbf{0.777} & \textbf{0.778} & 0.306 \\
\bottomrule
\end{tabular}
\label{tab:add_ablate_loss}
\end{table*}

\subsection{Compatibility with another one-step backbone}

As discussed in the main paper, SwiftPie can be further improved with stronger generative priors from more powerful one-step generative models. To demonstrate this, we adopt the same training strategy described in the main paper and train SwiftPie using an alternative one-step model, DMDv2 with an SDXL backbone. We report both qualitative and quantitative results in \cref{fig:add_sdxl_qual} and \cref{tab:add_sdxl_quan}. As shown in \cref{tab:add_sdxl_quan}, SwiftPie with the SDXL backbone achieves higher identity preservation and better text alignment score compared to SD1.5 backbone. Due to increase model's size, runtime increases from 0.17s to 0.41s; however, this remains significantly faster than multi-step approaches. In \cref{fig:add_sdxl_qual}, SwiftPie can generate personalized images with good generalization and strong alignment in both subject fidelity and text prompt.

\begin{table*}[ht]
\centering
\small
\setlength{\tabcolsep}{8pt}
\caption{Quantitative results of SwiftPie with different one-step backbone models on DreamBench}
\begin{tabular}{llcccccc} 
\toprule
\multirow{2}{*}{\textbf{Type}} & \multirow{2}{*}{\textbf{One-step Model}} & \multicolumn{3}{c}{\textbf{Subject Preservation}} & \multicolumn{1}{c}{\textbf{Text Alignment}} & \multirow{2}{*}{\textbf{Runtime} $\downarrow$} \\  
\cmidrule(lr){3-5}
\cmidrule(lr){6-6} 
& & CLIP-I $\uparrow$ & DINO $\uparrow$ & Nexus $\uparrow$ & CLIP-T $\uparrow$ & (seconds) \\
\midrule
\multirow{1.4}{*}{\begin{tabular}[c]{@{}l@{}}\textbf{SwiftPie}\\\end{tabular}} 
& SwiftBrushv2 \cite{dao2025swiftbrush} (SD1.5 backbone) & 0.862 & 0.777 & 0.778 & 0.306 & \textbf{0.17s} \\

& DMDv2 \cite{yin2024improved} (SDXL backbone) & \textbf{0.868} & \textbf{0.787} & \textbf{0.795} & \textbf{0.310} & 0.41s \\

\bottomrule
\end{tabular}
\vspace{-5pt}
\label{tab:add_sdxl_quan}
\vspace{-10pt}
\end{table*}


\section{User Study}
\label{sec:user_study}
\begin{figure}[H]
    \centering
    \includegraphics[width=.9\columnwidth]{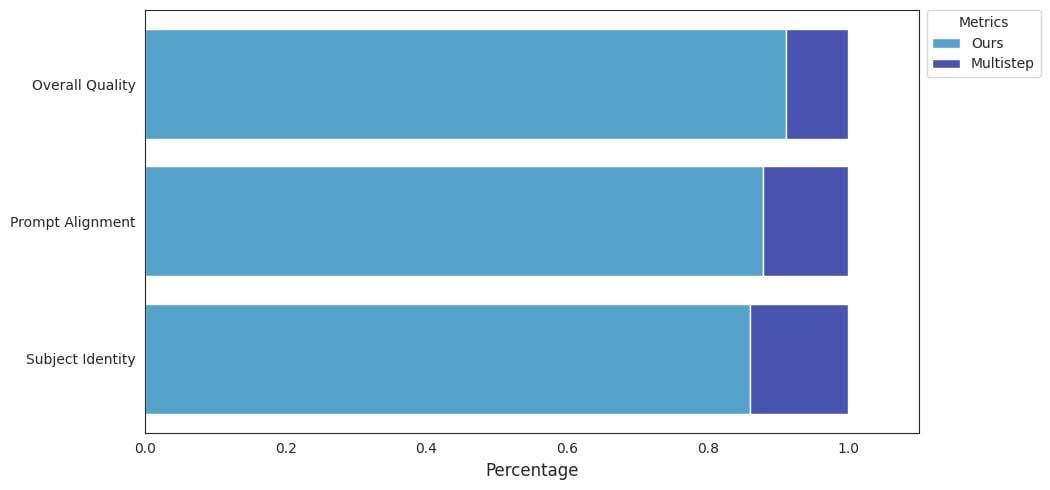}
    \caption{User study results}
    \label{fig:add_user_study}
\end{figure}

We conducted an additional user study to compare preferences between our one-step personalization results and those produced by other multi-step approaches. Specifically, we selected 5 subject identities and 4 prompts per subject from the DreamBooth dataset as inputs to both SwiftPie and several multi-step baselines. For each of the resulting 20 paired samples (one from SwiftPie and one from a multi-step method), approximately 20 participants rank their preferred output based on three criteria: identity preservation, prompt alignment, and overall image quality. As shown in \cref{fig:add_user_study}, SwiftPie emerges as the preferred method across all criteria, while also being the fastest: 86\% of users favored it for identity preservation, 87\% for prompt alignment, and 91\% for overall image quality.

\section{Societal Impact}
\label{sec:soc_impact}
SwiftPie is an AI-powered visual generation framework that can enable fast personalization with high-fidelity subject preservation and strong text alignment. Its exceptional speed makes it well-suited for interactive content creation tasks. However, SwiftPie also raises potential societal concerns, as it could be misused for unethical purposes, such as generating sensitive or harmful content to spread misinformation. This necessitates future research works on detecting and localizing AI-manipulated images to alleviate such potential issues.
